# Model-Based Diagnosis with Qualitative Temporal Uncertainty

**Wolfgang Nejdl**
Informatik V
RWTH Aachen
Ahornstraße 55, D-52056 Aachen
nejdl@informatik.rwth-aachen.de

**Johann Gamper**
Informatik V
RWTH Aachen
Ahornstraße 55, D-52056 Aachen
gamper@informatik.rwth-aachen.de

## Abstract

In this paper we describe a framework for model-based diagnosis of dynamic systems, which extends previous work in this field by using and expressing temporal uncertainty in the form of qualitative interval relations a la Allen. Based on a logical framework extended by qualitative and quantitative temporal constraints we show how to describe behavioral models (both consistency- and abductive-based), discuss how to use abstract observations and show how abstract temporal diagnoses are computed. This yields an expressive framework, which allows the representation of complex temporal behavior allowing us to represent temporal uncertainty. Due to its abstraction capabilities computation is made independent of the number of observations and time points in a temporal setting. An example of hepatitis diagnosis is used throughout the paper.

## 1 INTRODUCTION

Since most real world systems are dynamic, recently several extensions to the traditional model-based diagnosis approach have been developed with an explicit or implicit representation of time. Friedrich et al. propose [Friedrich and Lackinger, 1991] a very general extension of the traditional consistency-based approach to deal with temporal misbehavior. The dynamic behavior can be any set of First-Order sentences. The approaches in [Console et al., 1992; DeCoste, 1990; Downing, 1993] commonly approximate a dynamic system by a sequence of static systems, each of them can be modeled by the traditional static framework. The temporal reasoning framework in [Console and Torasso, 1991a] is based on a causal network, where time intervals are associated with both arcs (representing delays) and nodes (representing temporal extents).

In this paper we present an alternative framework for model-based diagnosis of dynamic systems by extend-ing the work in [Nejdl and Gamper, 1994]. Our main focus is the use of uncertainty in temporal diagnosis, by utilizing qualitative representations of complex temporal behavior and abstractions of observations from single time points into intervals. Additionally we include quantitative constraints on these intervals as well. This yields an expressive and efficient framework for diagnosis of time-varying systems.

In section 2 we introduce a hepatitis example, motivate our work and describe shortly the basic temporal framework. In section 3 we describe two different behavioral models, which are used for abductive and consistency based reasoning respectively. Section 4 discusses the concept of abstract observations, which makes diagnosis independent from time points. In section 5 we define explanation in our temporal framework and describe procedures for the generation of candidates. Finally, in section 6 we define abstract temporal diagnoses and show their computation.

## 2 PRELIMINARIES

### 2.1 EXAMPLE AND MOTIVATION

**Example 1** (Diagnosis of hepatitis A and B) In routine testing of the hepatitis A serology the findings *HAV*, *IgManti-HAV* and *anti-HAV*, in the case of hepatitis B the findings *HBsAg*, *anti-HBs*, *HBeAg*, *anti-HBe*, *anti-HBc* and *IgManti-HBc* are obtained, each of them can assume the value *positive* or *negative*. The natural course of a hepatitis infection is characterized by a typical sequence of findings [Horak and Adlassnig, 1990]: 1 variant for the hepatitis A, 4 acute and 4 persisting variants for the hepatitis B (figure 1). In each variant different stages can be distinguished: *no contact*, *incubation*, *acute*, *convalescence*, *immunity*.

The quantitative temporal relations in figure 1 are average values and they usually vary in each individual case. The qualitative relations among findings are much more reliable. All variants look similar, involving basically the same findings. What distinguishes these variants is the order in which these findings occur. The findings are constant over long time periods.

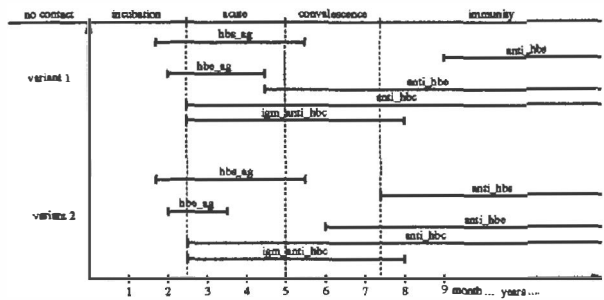

Figure 1: Sequence of positive findings in two acute variants of hepatitis B.

Therefore, the following properties are important for diagnosing such a dynamic system: First, describing behavior as a complex pattern of temporal interlocked symptoms. Second, explicit representation of qualitative and quantitative temporal relations. Third, abstraction mechanisms to reduce the complexity.

None of the current approaches provides all these facilities. In this paper we present a framework which extends model-based diagnosis into these directions. We extend our language with a subset of Allens interval algebra to describe dynamic behavior. We introduce two different behavioral models: the abductive model is used to generate explanation as covering, while the consistency constraint model must be satisfied by a diagnosis and is used to reduce the number of possible diagnoses and/or to strengthen the constraints used in the representation of diagnoses. The definition of abstract observations will lead to an abstraction of observations from time points to intervals. Finally we will define abstract temporal diagnoses as behavioral mode assumptions over time intervals which are described by means of qualitative and quantitative temporal relations.

## 2.2 BASIC TEMPORAL FRAMEWORK

Our basic temporal entities are time intervals (convex sets of time points). We use $t^-$ and $t^+$ to denote the start- and end-points of the interval $t$ respectively. If $t^-$ is equal to $t^+$ then $t$ represents a time point.

Allen's *Interval Algebra IA* [Allen, 1983] is based on a set $I = \{b, m, o, s, d, f, e, bi, mi, oi, si, di, fi\}$ of 13 basic mutually exclusive relations that can hold between two intervals. Indefinite knowledge is expressed as disjunction of basic relations and represented as a set. The IA provides a powerful framework to represent qualitative temporal information.

Van Beek [van Beek, 1991] defines the *Simple Interval Algebra SIA* as the subset of *IA*, which can be encoded entirely as conjunctions of continuous point relations $\{<, \leq, =, \geq, >, ?\}$ among the end-points of the intervals. We represent assertions in SIA as an *SIA-network*, where the nodes represent intervals and the arcs are labeled with the relation among the connected intervals. Van Beek gives tractable algorithms to answer interesting classes of queries in SIA. Given are a set of events $E$, a variable-free logical formula $\varphi$ involving temporal constraints between some of the events in $E$ and a SIA-network $C$ representing temporal information between the events in $E$. The algorithm *Possible* answers the query "Is $\varphi$ with respect to $C$ possibly true", i.e. is there at least one consistent instantiation of $C$ which satisfies also $\varphi$. The algorithm *Necessary* answers the query "Is $\varphi$ with respect to $C$ necessarily true", i.e. is $\varphi$ in each consistent instantiation of $C$ satisfied.

## 3   BEHAVIORAL MODEL

Given a system with components $COMPS$ we assume that each component has associated a set of *behavioral modes*. A component can assume a behavioral mode over an arbitrary time interval whose temporal extent is constrained relative to other intervals using qualitative and/or quantitative temporal relations.

**Definition 1** (ATBMA) An *Abstract Temporal Behavioral Mode Assumption* (ATBMA), stating that a component $c$ assumes behavioral mode $b$ during the time interval $t$, is defined as a formula

$$b(c, t) \wedge C(t)$$

where $C(t)$ is a set of *SIA*-relations and/or quantitative relations constraining the interval $t$.

The set $C(t)$ of qualitative and quantitative temporal constraints determines the *temporal extent* $t$ relative to other intervals[1]. This allows to represent indefinite knowledge about behavioral mode assumptions (i.e. "The onset of the disease occured during last week"). The behavior of the system is represented as the consequence of the behavioral modes, the *behavioral model*. In the following we will discuss two different behavioral models.

## 3.1   ABDUCTIVE BEHAVIORAL MODEL

**Definition 2** (Abductive Behavioral Model)
The *Abductive Behavioral Model $BM^+$* of a component $c \in COMPS$ assuming the behavioral mode $b$ over the time interval $t$ is defined as a formula

$$b(c, t) \wedge C(t) \rightarrow B_t \wedge B_s$$

where $B_t$ is a set of SIA-relations among manifestations over arbitrary time intervals and $B_s$ is a set of static constraints.

A *manifestation* $m(v, t)$ denotes the fact that the parameter $m$ assumes the value $v$ over time interval $t$. Using the definition above, the *temporal behavior* of a behavioral mode is described by a set $B_t$ of SIA-relations among manifestations. We represent consequences of a behavioral mode as an indefinite, complex pattern of manifestations over arbitrary time intervals. In $B_t$ we describe only the order in which the manifestations have to appear. This kind of qualitative temporal uncertainty is very important in medical domains as the absolut duration of manifestations usually varies in each individual case. Contrary to [Console *et al.*, 1992] we do not require that the manifestations arise at the same time as the behavioral mode. The set $B_s$ describes the *static behavior* and can be any other kind of constraints. In our example we have quantitative temporal constraints in $B_s$.

---

[1]As we will see later, these intervals usually are manifestations (of symptoms) which are connected to the real time line by the actual observations.

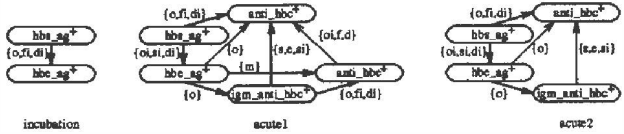

Figure 2: $SIA\text{-}BM^+$'s for *incubation*, *acute1* and *acute2*. The manifestations are denoted by intuitive abbreviations.

**Example 2** (Contd.) We model hepatitis using two components representing the two different types of virus infection, i.e. $COMPS = \{a, b\}$. In the following we will concentrate on the hepatitis B.

The different stages are modeled as behavioral modes (*incubation*, *acute1*, *acute2*, ...)[2]. The behavior of the stages is described as specific pattern of findings (*hbs_ag*, *anti_hbs*, *hbe_ag*, ...). The behavioral model $BM_{in}^+$ for the incubation stage is

$incubation(b, t_{in}) \land \{t_{in}\{o\}t_{hbs\_ag}, t_{in}\{o\}t_{hbe\_ag}\} \rightarrow$
$\{hbs\_ag(p, t_{hbs\_ag})\{o, fi, di\}hbe\_ag(p, t_{hbc\_ag})\} \land$
$\{t_{in} < 3\}$

The incubation stage is characterized by the appearance of positive *hbs_ag* and some time later of positive *hbe_ag*, represented by the relation $\{o, fi, di\}$ in $B_t$. The temporal extent $t_{in}$ overlaps (o-relation) the temporal extent of these findings. In $B_s$ we constrain $t_{in}$ to be less than 3 months.

The set $B_t$ of a behavioral model $BM^+$ can be represented as an SIA-network, which we denote as $SIA\text{-}BM^+$. The nodes represent the manifestations, the arcs are labeled with the SIA-relation among the connected manifestations. Figure 2 shows the $SIA\text{-}BM^+$ of the behavioral modes *incubation*, *acute1* and *acute2*. For clarity not all relations are shown.

The abductive behavioral model is used to generate explanations for observations. We model the faulty behavior using only the positive findings. Therefore we only want to explain positive observations. As this is not sufficient in all cases, we use additional consistency constraints to produce the correct diagnoses.

---

[2]Stages with the same manifestation pattern in different variants are modeled by a single behavioral mode, e.g. *incubation* represents the incubation stage in each variant.

## 3.2  CONSISTENCY CONSTRAINT MODEL

In our example, a set of positive observations corresponding to *acute2* can be covered by *acute1* as well as by *acute2*. To avoid this effect we require that no observation is contradictory with the predicted manifestations. Moreover, the incubation stage behavioral model (only predicting positive manifestations) covers also the positive *hbs_ag* and *hbe_ag* of the acute stage. Applying a closed world assumption yields the constraint that all other manifestations have to be negative (in particular *anti_hbc*) and therefore ends the incubation stage as soon as *anti_hbc* is positive.

**Definition 3** (Consistency Constraint Model) The *Consistency Constraint Model* $BM^-$ for an abductive behavioral model $BM^+$ is a formula $b(c, t) \land C(t) \rightarrow B_t \land B_s$ and is defined as

$$BM^- = Nec(BM^+, \Sigma^+) \uplus CWA(BM^+, \Sigma^-)$$

where $\Sigma^+$ and $\Sigma^-$ are two sets of manifestations and $\uplus$ denotes the union of two behavioral models.

$Nec(BM^+, \Sigma^+)$ determines that part of the abductive behavioral model, which has to be present in all cases. The closed world assumption $CWA(BM^+, \Sigma^-)$ states that all findings in $\Sigma^-$ which are not used in $BM^+$ have to be negative.

**Example 3** (Contd.) We have $\Sigma^+ = \Sigma^- = \{hbs\_ag,$ *anti_hbs*, *hbe_ag*, *anti_hbe*, *anti_hbc*, *igm_anti_hbc*$\}$. Hence, the necessary part is always the same as the abductive behavioral model. The consistency constraint model $BM_{in}^-$ for the incubation stage is computed as

$incubation(b, t_{in}) \land \{t_{in}\{o\}t_{hbs\_ag}, t_{in}\{o\}t_{hbe\_ag},$
$t_{in}\{d\}t_{anti\_hbs}, t_{in}\{d\}t_{anti\_hbe},$
$t_{in}\{d\}t_{anti\_hbc}, t_{in}\{d\}t_{igm\_anti\_hbc}\} \rightarrow$
$\{hbs\_ag(p, t_{hbs\_ag})\{o, fi, di\}hbe\_ag(p, t_{hbe\_ag}),$
$anti\_hbs(n, t_{anti\_hbs})\{cont\}anti\_hbe(n, t_{anti\_hbe}), \ldots\}$

The necessary part is the same as the abductive behavioral model shown in the last example. The closed world assumption states that in the incubation stage the findings *anti_hbs*, *anti_hbe*, *anti_hbc* and *igm_anti_hbc* have to be negative. This is expressed by a d-relation (during) among $t_{in}$ and the temporal extent of each of these findings. Obviously, these findings have a common subinterval represented by a *cont*-relation (contemporary), defined as $I \setminus \{b, m, bi, di\}$, among each pair of them.

If a single ATBMA cannot explain all observations, we get an inconsistency with the CWA. Obviously, by combining different behavioral modes we decrease the closed world part of the consistency constraint model and get the corresponding multiple faults.

# 4   OBSERVATIONS AND ABSTRACT OBSERVATIONS

An *observation* is a measurement of a parameter at a time point. We write $obs(v, t)$ to denote that for parameter $obs$ we measured the value $v$ at time $t$. $OBS(t)$ is the set of all observations in the time interval $t$.

In many applications, such as in the hepatitis example, we assume a continuous persistence of parameters: given an observation $obs(v, t)$, the parameter has value $v$ at time point $t$ and possibly the same value before and after $t$.

**Definition 4** (Abstract Observation)
An *Abstract Observation* for parameter $aobs$ assuming value $v$ over the time interval $t$ is defined as

$$aobs(v, t) \land C(t) \land f : aobs \to \{obs_i\}$$

where $C(t)$ is a set of SIA-relations among the temporal extent $t$ and intervals on the real time line and $f$ is a partial function, which defines a set of covered observations $obs_i$ for parameter $aobs$, such that the value of each $obs_i$ is $v$ and the observation time point is in interval $t$. We will leave out $f$ whenever it is clear from the context.

An abstract observation represents the assumption that a parameter has a value over a time interval. We cannot determine exactly its temporal extent, but we constrain it relative to the covered observations using the qualitative temporal relations in SIA. An abstract observation which covers as many as possible consecutive observations with the same value is called *maximal*. If it is not explicitly mentioned we always use maximal abstract observations.

The concept of abstract observations is an important shift from a discrete view based on time points to a view driven by changes of observations independent from the granularity of time [Ginsberg, 1991]. This can improve considerably the diagnostic process in the case when parameters are stable over long time periods.

**Example 4** (Cont.) We assume the observations of all 6 hepatitis B findings shown in figure 3a. For the positive *hbs_ag* we construct the maximal abstract observation $hbs\_ag(p, t) \land \{t\{oi\}[1, 2], t\{o\}[5, 6]\} \land \{hbs\_ag(p, 2), hbs\_ag(p, 3), hbs\_ag(p, 4), hbs\_ag(p, 5)\}$ indicating that the interval over which we assume that $hbs\_ag$ is positive starts between 1 and 2 and ends between 5 and 6. We have 6 maximal abstract observations for the positive findings. The remaining 5 are:

$$
\begin{aligned}
anti\_hbs(p, t) \quad &\land \quad \{t\{oi\}[7, 8], t\{di\}[8, 9]\} \\
hbe\_ag(p, t) \quad &\land \quad \{t\{oi\}[1, 2], t\{o\}[3, 4]\} \\
anti\_hbe(p, t) \quad &\land \quad \{t\{oi\}[5, 6], t\{di\}[6, 9]\} \\
anti\_hbc(p, t) \quad &\land \quad \{t\{oi\}[2, 3], t\{di\}[3, 9]\} \\
igm\_anti\_hbc(p, t) \quad &\land \quad \{t\{oi\}[2, 3], t\{o\}[7, 8]\}
\end{aligned}
$$

We denote a set of abstract observations constructed from a set $OBS(t)$ as $AOBS(t)$. A set $AOBS(t)$ and the qualitative temporal relations among them can be represented as an SIA-network, called $SIA\text{-}AOBS$. Figure 3 shows the SIA-AOBS representing the positive abstract observations in our example.

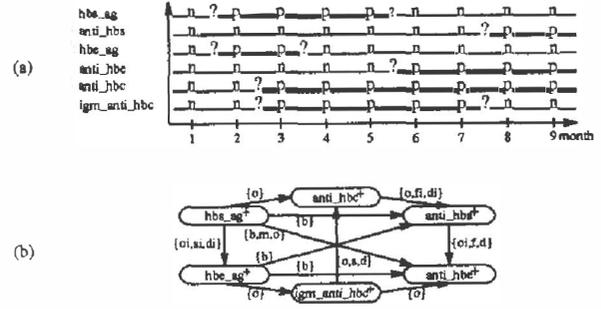

Figure 3: (a) Observations in our example. A "p" ("n") denotes the measurement of a positive (negative) value, thick (thin) lines indicate the temporal extent of positive (negative) abstract observations. (b) $SIA\text{-}AOBS^+$ representing the positive abstract observations denoted by intuitive abbreviations.

# 5   CANDIDATE GENERATION

For the calculation of an abstract temporal diagnosis we have to construct an explanation for a given set of observations at arbitrary time points as well as the temporal relations among them. Similar to [Console and Torasso, 1991b] we propose an abductive approach with additional consistency constraints. The set $OBS^+$ of observations that has to be covered by a diagnosis is the set of positive findings (abnormality observations), the set $OBS^-$ used for consistency checking is the set of all observations.

In this section we discuss the generation of candidates, which is an important step in the computation of abstract temporal diagnoses. In particular, we show how the different use of the abductive behavioral model and the consistency constraint model, how the use of abstract observations instead of observations leads to efficient algorithms, and finally, the evaluation of the static constraints.

A *candidate* is a tuple $\langle ATBMA, CAOBS^+ \rangle$ where $ATBMA$ is an abstract temporal behavioral mode assumption which covers the nonempty set $CAOBS^+$ of abstract observations.

Procedure *Candidates* (figure 4) takes as input a behavioral model $BM = \langle BM^+, BM^- \rangle$ and a set $AOBS = \langle AOBS^+, AOBS^- \rangle$ of abstract observations. According to the two different behavioral models we generate abductively a hypothesis $h$ (hypothetical candidate), which is tested against the consistency constraints resulting in $h'$ and against the static constraints resulting in $h''$. Both of these tests may lead to a hypothesis with tighter constraints or even an inconsistent hypothesis. A consistent hypothesis $h''$ represents a candidate and is added to the set $C$. We repeat this process until no new hypotheses can be generated (step 4). Candidates returns the set of all candidates for the behavioral model $BM$ and the abstract observations $AOBS$.

**Algorithm 1** *Candidates(BM, AOBS)*
- $C = \emptyset$
- **loop forever**
- $\quad h = Abduction(BM^+, AOBS^+)$
- $\quad$ **if** $h = \langle \rangle$ **then return** $C$
- $\quad h' = Consistency(h, BM^-, AOBS^-)$
- $\quad h'' = Static(h', BM^+)$
- $\quad$ **if** $h''$ is consistent **then** $C \leftarrow C \cup h''$

Figure 4: Procedure Candidates.

## 5.1   ABDUCTION

Abduction uses explanation as covering which in our framework is characterized by the following definition.

**Definition 5** (Temporal Covering Condition) Given is a behavioral model $BM = b(c, t) \wedge C(t) \rightarrow B_t \wedge B_s$. The ATBMA $b(c, t) \wedge C(t)$ *covers* a set $AOBS$ of abstract observations iff the $SIA\text{-}AOBS$ is necessarily true in the $SIA\text{-}BM$ and $B_s$ is satisfied.

An SIA-network $G_1$ is necessarily true in an SIA-network $G_2$ iff each consistent instantiation of $G_2$ satisfies also $G_1$.

Procedure *Abduction* (figure 5) takes as input an abductive behavioral model $BM^+$ and a set $AOBS^+$ of abstract observations which has to be covered. Usually $AOBS^+$ is very large and cannot be covered by a single behavioral model. Thus, in step 1 we build a subset $CAOBS^+$ which contains an abstract observation for each corresponding manifestation in the temporal behavior $B_t$ of $BM^+$. Then we test the temporal covering condition (definition 5) by invoking procedure Necessary [van Beek, 1991] with $SIA\text{-}BM^+$ and $SIA\text{-}CAOBS^+$ as parameters[3]. If Necessary succeeds the abstract observations in $CAOBS^+$ and the temporal relations among them are covered by $BM^+$. We invoke procedure Instantiate to generate an ATBMA, which together with the set $CAOBS^+$ is returned as an abductive hypothesis. If Necessary fails we invoke procedure Splitting to construct new abstract observations $CAOBS^+_{new}$ with smaller temporal extents. If $CAOBS^+_{new}$ is the empty set no further splitting is possible and we consider a new subset of $AOBS^+$. Procedure Abduction returns an empty tuple if no hypothesis could be generated.

[3]Necessary only works on $B_t$, $B_s$ is tested in a later step.

**Algorithm 2** *Abduction(BM$^+$, AOBS$^+$)*
- **while** get next $CAOBS^+_{new} \subseteq AOBS^+$ **do**
- $\quad$ **repeat**
- $\quad\quad CAOBS^+ \leftarrow CAOBS^+_{new}$
- $\quad\quad$ **if** $Necessary(SIA\text{-}BM^+, SIA\text{-}CAOBS^+)$ **then**
- $\quad\quad\quad ATBMA \leftarrow Instantiate(BM^+, CAOBS^+)$
- $\quad\quad\quad$ **return** $\langle ATBMA, CAOBS^+ \rangle$
- $\quad\quad$ **else**
- $\quad\quad\quad CAOBS^+_{new} \leftarrow Splitting(CAOBS^+, SIA\text{-}BM^+)$
- $\quad$ **until** $CAOBS^+_{new} = \emptyset$
- **return** $\langle \rangle$

Figure 5: Procedure Abduction.

### 5.1.1   Splitting Abstract Observations

In the construction of abstract observations in section 4 we had no knowledge about the causes of observations. Thus, it might turn out that an abstract observation is caused by several behavioral modes. If a set of abstract observations violates the temporal covering condition we construct new abstract observations with smaller temporal extents.

**Definition 6** (Splitting Abstract Observations) The result of *splitting* an abstract observation $aobs1 = aobs(v, t_1) \wedge C(t_1) \wedge f_1$ is a new abstract observation $aobs2 = aobs(v, t_2) \wedge C(t_2) \wedge f_2$, such that

- $t_2\{s, d, f\}t_1$
- $f_2(aobs2) \subset f_1(aobs1)$

Splitting an abstract observation $aobs1$ removes at least one of the covered observations and the temporal extent gets smaller.

In step 8 in procedure Abduction the temporal covering condition for $CAOBS^+$ is violated. Thus, we try to split (some of) these abstract observations, which produces modified temporal relations. Procedure *Splitting* returns such a new set $CAOBS^+_{new}$ of abstract observations. The discrepancies between the temporal behavior $SIA\text{-}BM^+$ and the old $SIA\text{-}CAOBS^+$ are used to improve the splitting process. We never split an abstract observation if all relations in which it appears are satisfied. Further, we split all abstract observations violating the temporal covering condition at once. Since each call of Splitting shortens at least one of the abstract observations[4], the test and splitting loop in Abduction will terminate in any case.

**Example 5** (Contd.) The set $AOBS^+$ to be covered by a diagnosis is the set of positive abstract observations. The $SIA\text{-}AOBS^+$ is shown in figure 3b.

Let us consider the incubation stage. The subnetwork of $SIA\text{-}AOBS^+$ in question consists of $hbs\_ag^+$ and $hbe\_ag^+$ and the arc among them labeled with

[4]Splitting an abstract observation which covers exactly one observation corresponds to removing it.

$\{oi, si, di\}$. This subnetwork is not necessarily true in $SIA\text{-}BM_{in}^+$ (because of the $o$- and the $fi$-relation). Therefore we split $hbs\_ag^+$ to $hbs\_ag^{+\prime}$, which covers only the positive $hbs\_ag$ at time 2 and 3 and which has the constraints $\{t_{hbs\_ag^{+\prime}}\{oi\}[1,2]$, $t_{hbs\_ag^{+\prime}}\{o\}[3,4]\}$. The relation among $hbs\_ag^{+\prime}$ and $hbe\_ag^+$ is $\{cont\}$ (contemporary) and is necessarily true in the $SIA\text{-}BM_{in}^+$. Thus, the incubation stage covers the abstract observations $hbs\_ag^{+\prime}$ and $hbe\_ag^+$.

In this example we have seen how the use of abstract observations leads to an event-driven [Ginsberg, 1991] reasoning independent from the number of specific observations and granularity of time. The $4 \times 2 = 8$ possible tuples of positive observations for $hbs\_ag$ and $hbe\_ag$ have been reduced to a single tuple of corresponding abstract observations. The systems in [Console et al., 1992; Downing, 1993] would perform diagnosis at each time point. The gain of efficiency in our framework depends highly on the frequency and persistency of observations.

### 5.1.2 Instantiate

In step 5 in Abduction we have found a set $CAOBS^+$ of abstract observations covered by $BM^+$ and we invoke procedure *Instantiate* to generate an ATBMA. After instantiating the manifestations in $BM^+$ to the abstract observations in $CAOBS^+$ procedure Instantiate evaluates the union of the following relations: the relations in $C(t)$ that constrain the temporal extent $t$ of the behavioral mode relative to the manifestations in $B_t$ and the relations from the abstract observations $CAOBS^+$ which constrain them relative to the real time line. Evaluating these qualitative constraints corresponds to finding all feasible relations, which in SIA can be solved by a polynomial algorithm [van Beek, 1991]. This leads to a description of the temporal extent $t$ over which component $c$ assumes behavioral mode $b$ in terms of qualitative temporal SIA-relations.

**Example 6** (Contd.) We instantiate the manifestations in $BM_{in}^+$ to the abstract observations $hbs\_ag^{+\prime}$ and $hbe\_ag^+$. Evaluating the constraints $\{t_{in}\{o\}t_{hbs\_ag^{+\prime}},\ t_{in}\{o\}t_{hbe\_ag^+}\} \cup \{t_{hbs\_ag^{+\prime}}\{oi\}[1,2],$ $t_{hbs\_ag^{+\prime}}\{o\}[3,4],\ t_{hbe\_ag^+}\{oi\}[1,2],\ t_{hbe\_ag^+}\{o\}[3,4]\}$ (from $BM_{in}^+$ and $CAOBS^+$ respectively) leads to

$incubation(b, t_{in}) \wedge \{t_{in}\{cont\}[1,2], t_{in}\{b,m,o\}[3,4]\}$

stating that the incubation stage is present some time between 1 and 2 and ends before time 4. This ATBMA together with $\{hbs\_ag^{+\prime}, hbe\_ag^+\}$ is a hypothesis for the incubation stage.

### 5.2 CONSISTENCY

In this section we will show, how the consistency constraint model might lead to tighter constraints for abductively generated hypotheses. We start with the definition of explanation as consistency in our framework.

**Definition 7** (Temporal Consistency Condition)
Given is a behavioral model $BM = b(c,t) \wedge C(t) \rightarrow B_t \wedge B_s$. The ATBMA $b(c,t) \wedge C(t)$ is *consistent* with a set $AOBS$ of abstract observations iff the $SIA\text{-}AOBS$ is possibly true in the $SIA\text{-}BM$ and $B_s$ is satisfied.

An SIA-network $G_1$ is possibly true in an SIA-network $G_2$ iff there is at least one consistent instantiation of $G_2$ which satisfies $G_1$.

In step 5 in Candidates we invoke procedure *Consistency*, which tests the abductively generated hypothesis $h = \langle ATBMA, CAOBS \rangle$ against the consistency constraint model $BM^-$ plus the set $AOBS^-$ of all abstract observations. Consistency works similar to Abduction, except for using procedure Possible [van Beek, 1991] to test the temporal consistency condition. It generates an ATBMA for $BM^-$ by using the positive abstract observations in $CAOBS^+$ plus negative abstract observations from $AOBS^-$, which has to be consistent with the ATBMA in $h$. This leads to a new ATBMA which might have tighter constraints, and together with the accordingly modified set $CAOBS^+$ is returned as new hypothesis $h'$.

**Example 7** (Contd.) We test whether the ATBMA for the incubation stage generated in the abductive step is consistent with $BM_{in}^-$. Considering only the negative *anti_hbc* and evaluating the constraints $\{t_{in}\{d\}t_{anti\_hbc^-}\} \cup \{t_{anti\_hbc^-}\{di\}[1,2],$ $t_{anti\_hbc^-}\{o\}[2,3]\}$ (from $BM_{in}^-$ and $AOBS^-$) leads to $incubation(b, t_{in}) \wedge \{t_{in}\{b,m,o,d,s\}[2,3]\}$, which is still consistent with the old hypothesis. As the new constraints are tighter we get the new ATBMA

$incubation(b, t_{in}) \wedge \{t_{in}\{cont\}[1,2], t_{in}\{o\}[2,3]\}$

### 5.3 EVALUATING THE STATIC CONSTRAINTS

So far we considered only the temporal behavior and generated a hypothesis satisfying the temporal covering and consistency conditions respectively. The last step in the generation of candidates is to test the static constraints (step 5 in Candidates). Usually, this can be any kind of constraints and we can exploit the traditional model-based diagnosis framework to evaluate them. In our example we have only quantitative temporal constraints. In the following we will show how these constraints might lead to tighter constraints for the hypothesis generated so far.

We discuss only the case where the maximal temporal extent $t$ of an ATBMA is constrained, i.e. $B_s$ contains constraints of the form $t < d$, and $d$ is a real number. The set $C(t)$ of an ATBMA can be considered to be of the form $\{t\{oi\}s, t\{o\}e\}$, where $s = [s^-, s^+]$ is the interval in which $t$ starts, $e = [e^-, e^+]$ is the interval in which $t$ ends, and $s$ and $t$ do not overlap each other[5].

---

[5] Such a representation (similar to a variable interval in [Console and Torasso, 1991a]) is always possible, since we impose that an ATBMA covers at least one observation.

We can distinguish 4 cases:

1. $e^- - s^+ > d$: $C(t)$ is inconsistent.

2. $e^+ - s^- < d$: $t < d$ is already satisfied.

3. $(e^- - s^+ < d) \wedge (e^+ - s^+ > d)$: Set $e^+ \leftarrow s^+ + d$ and add $t < d$ to $C(t)$.

4. $(e^- - s^+ < d) \wedge (e^- - s^- > d)$: Set $s^- \leftarrow e^- - d$ and add $t < d$ to $C(t)$.

In case 1 we reject the hypothesis. In all other cases, the set $C(t)$ is consistent, in case 3 and 4 we get tighter constraints. Note that in this case we have to modify the set of covered abstract observations accordingly.

**Example 8** (Contd.) We translate the temporal constraints in the hypothesis for the incubation stage to $\{t_{in}\{oi\}[-\infty, 2], t_{in}\{o\}[2,3]\}$. In the set $B_s$ of the incubation stage we have the quantitative temporal constraint $t_{in} < 3$. Evaluating this constraint (case 4) leads to the modified ATBMA

$incubation(b, t_{in}) \wedge \{t_{in}\{oi\}[-1, 2], t_{in}\{o\}[2, 3], t_{in} < 3\}$

In other cases, we need the quantitative constraints to distinguish between acute and persisting variants of hepatitis B.

# 6  ABSTRACT TEMPORAL DIAGNOSES

**Definition 8** (Associated observations)
The set $OBS$ of observations *associated* to a set $AOBS$ of abstract observations is defined as $OBS = \bigcup_{aobs \in AOBS} f(aobs)$.

In many cases temporal constraints between different behavioral modes are known. We represent them in a *Mode Constraint Graph SIA-MC*, where the nodes represent behavioral modes and the arcs are labeled with the allowed SIA-relation among the connected modes (see [Nejdl and Gamper, 1994] for details). The system in [Portinale, 1992] assumes that probabilistic knowledge is available about the evolution of a system and represents the transition of behavioral modes by means of Markov Chains.

An abstract temporal diagnosis is an explanation for a set of observations at arbitrary time points as well as for the temporal relations among them. Remember, that we use abductive diagnosis with the set $OBS^+$ and consistency-based diagnosis with the set $OBS^-$.

**Definition 9** (Abstract Temporal Diagnosis)
Let $\mathcal{BM}^+$ be the set of all abductive behavioral models $BM^+$, $\mathcal{BM}^-$ the set of all consistency constraint models $BM^-$. An *Abstract Temporal Diagnosis* $D(t)$ is defined as

$$D(t) = W(t) \cup P$$

where $W(t)$ is a set of ATBMA's an $P$ is a path in $SIA\text{-}MC$ such that

**Algorithm 3** $ATD(\mathcal{BM}, SIA\text{-}MC, OBS)$

- $\mathcal{D} \leftarrow \emptyset$
- for each path $P$ from $SIA\text{-}MC$ do
- $\quad COBS^+_{new} \leftarrow OBS^+, D \leftarrow \emptyset$
- $\quad$ repeat
- $\quad\quad COBS^+ \leftarrow COBS^+_{new}$
- $\quad\quad$ get next $BM$ from $\mathcal{BM}$ (according $P$)
- $\quad\quad C \leftarrow Candidates(BM, AOBS)$
- $\quad\quad$ for each $\langle ATBMA, CAOBS^+ \rangle \in C$ do
- $\quad\quad\quad$ if $ATBMA$ is consistent with $P$ then
- $\quad\quad\quad\quad COBS^+_{new} \leftarrow COBS^+ \setminus \bigcup_{aobs \in CAOBS^+} f(aobs)$
- $\quad\quad\quad\quad D \leftarrow D \cup ATBMA$
- $\quad$ until $COBS^+_{new} = \emptyset$ or there is no new $BM$ in $P$
- $\quad$ if $COBS^+ = \emptyset$ then $\mathcal{D} = \mathcal{D} \cup (D \cup P)$
- return $\mathcal{D}$

Figure 6: Procedure ATD.

- $W(t)$ covers $OBS^+(t)$, i.e. $W(t) \cup \mathcal{BM}^+ \models OBS^+(t)$.
- $W(t)$ is consistent with $OBS^-(t)$, i.e. $W(t) \cup \mathcal{BM}^- \cup OBS^-(t)$ is consistent.
- $W(t)$ is consistent with the path $P$ in $SIA\text{-}MC$.

The following procedure $ATD$ takes as input parameters a set $\mathcal{BM}$ of behavioral models, a mode constraint graph $SIA\text{-}MC$ and a set $OBS$ of observations and returns the set of all abstract temporal diagnoses. The main idea of ATD is to generate candidates, which are consistent with a path in the mode constraint graph, until all observations in $OBS^+$ are explained. We first choose a path $P$ from the mode constraint graph and initialize the set $COBS^+_{new}$ to the observations which have to be covered. Then we get the next behavioral model $BM$ according to $P$ avoiding in this way to consider behavioral modes not appearing in $P$. We invoke procedure Candidates, which returns the set $C$ of all candidates for $BM$. Subsequently each of them is tested against $P$. If the $ATBMA$ of a candidate is consistent with $P$ we remove the associated observations from $COBS^+$ and add the $ATBMA$ to the set $D$. A set $D$ of ATBMA's which covers the whole set $OBS^+$ together with $P$ is a single abstract temporal diagnosis and we add it to the set $\mathcal{D}$ of all diagnoses.

**Example 9** (Example with sparse observations) We assume observations of all 6 findings at time 2 and 6 taken from figure 3.

For the incubation stage we generate an ATBMA with the constraints $C(t_{in}) = \{t_{in}\{o\}[-1, 2], t_{in}\{o\}[2, 5], t_{in} < 3\}$ covering the positive *hbs_ag* and *hbe_ag*.

Next we consider *acute2* and generate the ATBMA with $C(t_{ac2}) = \{t_{ac2}\{d\}[2, 6]\}$, which is consistent, but covers no abstract observation. Thus, we have no candidates for *acute2*.

For the next stage *convalescence2* we generate abductively the ATBMA with $\{t_{co2}\{bi, mi, oi, d, f\}[2, 6]\}$.

Evaluating the consistency constraint model leads to $\{t_{co2}\{oi\}[2,6]\}$, which still covers the remaining abstract observations and is consistent.

The candidates *incubation* and *convalescence2* are consistent with the $b$-relation in the mode constraint graph and we get the abstract temporal diagnosis

$\{$ *incubation*$(b, t_{in}) \wedge$
    $\{t_{in}\{oi\}[-1,2], t_{in}\{o\}[2,5], t_{in} < 3\},$
  *convalescence2*$(b, t_{co2}) \wedge \{t_{co2}\{oi\}[2,6]\},$
  $t_{in}\{b\}t_{co2} \}$

Due to the $b$-relation between $t_{in}$ and $t_{co2}$ we do not cover the whole interval during which observations are made. If we want to cover these time points we can add consistent ATBMA's to the diagnosis according the mode constraint graph. In particular, in this example we can add to the above diagnosis

$\{$ *acute2*$(b, t_{ac2}) \wedge \{t_{ac2}\{d\}[2,6]\},$
  $t_{in}\{m\}t_{ac2}, t_{ac2}\{m\}t_{co2} \}$

Abstract temporal diagnoses as sets of behavioral mode assumptions over indefinite time intervals represent in a natural way vague knowledge human diagnosticians often have about the evolution of a system. Different to the systems in [Console *et al.*, 1992; Downing, 1993; Portinale, 1992] this representation is independent of the granularity of time.

## 7    CONCLUSION

We proposed a framework for model-based diagnosis of dynamic systems, which extends previous work in this field in several ways. The use of qualitative temporal constraints a la Allen as well as quantitative temporal constraints considerably improve the expressiveness at the knowledge representation level. We describe dynamic behavior as complex pattern of manifestations, which are present over arbitrary time intervals. The automatically generated consistency constraint model leads to a more exact description of behavioral modes. The concept of abstract observations provides a change driven computation instead of a time point driven one and improves expressiveness as well as efficiency. Generating explanations in such a framework leads to abstract temporal diagnoses defined as behavioral mode assumptions over time intervals described in terms of qualitative and quantitative temporal relations.

We further use Allens interval algebra in a uniform way for behavioral modes, (abstract) observations, knowledge about behavioral modes and abstract temporal diagnoses themselves. This gives us a simple representation of qualitative and quantitative temporal uncertainty at different levels.